# Interpretable Feature Engineering for Time Series Predictors using Attention Networks

Tianjie Wang, Jie Chen, Joel Vaughan, and Vijayan N. Nair

Corporate Model Risk, Wells Fargo

May, 2022

## Abstract

Regression problems with time-series predictors are common in banking and many other areas of application. In this paper, we use multi-head attention networks to develop interpretable features and use them to achieve good predictive performance. The customized attention layer explicitly uses multiplicative interactions and builds feature-engineering heads that capture temporal dynamics in a parsimonious manner. Convolutional layers are used to combine multivariate time series. We also discuss methods for handling static covariates in the modeling process. Visualization and explanation tools are used to interpret the results and explain the relationship between the inputs and the extracted features. Both simulation and real dataset are used to illustrate the usefulness of the methodology.

## 1. Introduction

We consider regression problems with multivariate time series as predictors. Denote the time series as

$$X_{j,k}^{[i]} = X_j^{[i]}(t_k), j = 1, \ldots, m; \ k = 0, \ldots, T; \ i = 1, \ldots, n. \quad Eq\ (1)$$

Here $j = 1, \ldots, m$ are $m$ (multivariate) time series, $k = 0, \ldots, T$ are observation times, and $i = 1, \ldots, n$ are the multiple samples. We focus on modeling the responses $Y_i(T)$, at some fixed time $T$, as a function of the time-series predictors $\{X_{j,k}^{[i]}\}$. The methodology can be readily extended to modeling a sequence of outputs $\{Y_i(T - t_0), \ldots Y_i(T)\}$ one at a time, to forecasting over a horizon, etc.,

There is a huge literature on time-series modeling and forecasting in the statistical, econometric, finance, and related literature [1, 2]. There are also more recent approaches in the machine learning (ML) literature using recurrent neural networks (RNNs), long short-term memory networks (LSTMs) and other deep neural networks (DNNs). While they have good predictive performance, the results are hard to interpret. A more indirect approach is to do manual feature engineering of the time series and use these features as predictors in ensemble tree algorithms (such as GBM) or feedforward neural networks (FFNNs). However, this requires considerable subject matter expertise and is also time consuming. This paper studies alternative ML approaches that do automatic feature extraction, yield interpretable features, and result in effective modeling and forecasting.

Our methodology is based on the notion of attention mechanisms that have gained popularity in deep learning [3]. These developments were inspired by notable achievements of the Transformer architectures in natural language processing (NLP) applications [4]. Recent work has applied attention mechanisms in time series forecasting applications, with improved performance over comparable recurrent networks [5, 6, 7]. Our approach incorporates a multiplicative interaction in constructing the attention subnetwork, which allows for more parsimonious representation than feedforward neural networks (FFNNs). In addition, the generated features are quite interpretable. Unlike existing automatic time-series feature extraction methods, such as *tsfresh* [8] and TSFEL [9] which select from a feature library of univariate time series, the proposed architecture adapts to the datasets and can capture interactions across multivariate time series.

The remainder of the paper is organized as follows: Section 2 provides the structural components of the proposed methods of attention mechanism, building from univariate time series to multivariate time series. Section 3 presents simulation studies to illustrate the method under different applications and scenarios, including scenarios with static covariates. Then, Section 4 gives examples of real data analysis. Finally, the discussion, conclusion and suggested future work are included in Section 5.

## 2. Methodology: Attention and Heads

Attention mechanisms were originally developed for neural machine translation [3]. In this application, the inputs (words) were treated as a sequence and, conditional on a context vector, the model predicts the probability of the next word in the translation. In the sections below, we describe how the methodology can be adapted to tabular data with time series predictors.

### 2.1 Attention Layer for a Univariate Time Series

We start with a univariate time series $X^{[i]} = \left(X_0^{[i]}, \dots, X_T^{[i]}\right)$, with $i = 1, \dots, n$ samples. The atomic structure of the architecture is the attention layer, defined as

$$Feat^{[i]} = \sum_{k=0}^{T} A_k\left(X^{[i]}\right) s_k X_k^{[i]}, \qquad Eq\ (2)$$

where the $A_k\left(X^{[i]}\right)'s$ are "attention scores", and $s_k's$ are trainable scaling coefficients.

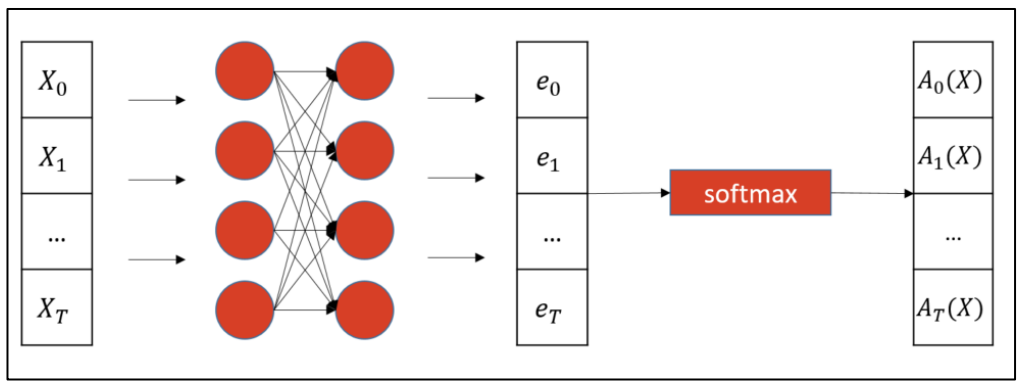

*Figure 1: Construction of Attention Subnet $A_k(X)$ from feed-forward subnets with softmax activation*

Figure 1 provides an illustration of this structure (called Attention Subnets), and it works as follows. For each sample $i = 1, \dots, n$, the time series input $(X_0, \dots, X_T)$ is first transformed into outputs $(e_0, , \dots, e_T)$ using a "small" FFNN. The softmax function

$$\frac{\exp(e_k)}{\sum_{\ell=0}^{T}\exp(e_\ell)}$$

is then applied to get $A_k = A_k(X^{[i]}), k = 0, \ldots, T$. The softmax property implies that $A_k(X^{[i]}) > 0$ and $\sum_{k=0}^{T} A_k(X^{[i]}) = 1$. The scaling constants $\{s_k\}$ in Eq (2) allow for different signs and hence more generality.

We see from Eq (2) that the feature is a weighted sum of the input time series $\left(X_0^{[i]}, \ldots, X_T^{[i]}\right)$, and the end result reflects the importance of time point $k$ in the feature. When $T$ is large, the softmax function forces some of the $A_k(X^{[i]})'s$ to be close to 0. This is like an instance-wise variable selection mechanism that decides which time points get more or less weight in the generated features. There has been recent work on sparse alternatives to softmax such as *sparsemax* that can force some $A_j(X^{[i]})$ to shrink to 0, which can further increases the variable selection strength [10]. The multiplicative interaction term $A_k(X^{[i]}) X_k^{[i]}$ in Eq (2) increases the expressivity of the model greatly (see [11]). This allows one to use a parsimonious network to learn more flexible features than FFNNs with similar numbers of trained parameters.

We will discuss interpretability of the features in later sections. To explain it here briefly, write $A_k(X^{[i]}) s_k X_k^{[i]} = W_k(X^{[i]})X_k^{[i]}$. We can then examine the generated feature $Feat^{[i]}$ in Eq (2) across all samples and examine the component(s) with large impact. By plotting the weight $W_k(X^{[i]})$ for each sample $i$, we can see the selection of inputs for each instance, which aids in interpreting the results.

## 2.2 Multivariate Time Series with Attention Heads

A naïve way to handle multivariate time series is to concatenate all the inputs into a one-dimensional array and apply the attention layer described in Eq (2). However, the flattening will destroy the multivariate time series structure and lose useful information from the time-dependent and cross-variable relationships. In this paper, we propose the combined use of convolution kernels and attention mechanism to handle the multivariate problem. Convolutional neural networks (CNNs) have been widely used in financial time series to capture the time structure across multiple variables [12].

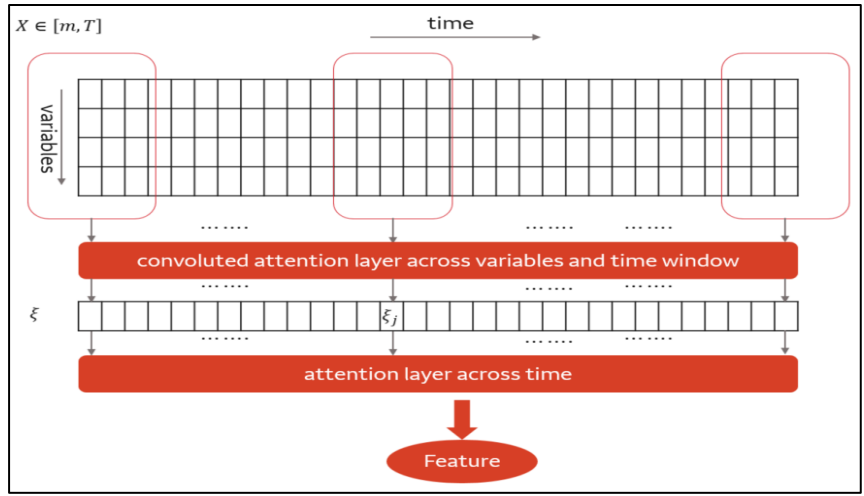

*Figure 2: Feature engineering head for multi-variate time series*

As shown in Figure 2, we use a shared-weight architecture of convolution kernels that slide along the time index. The weights of the kernels are changed from trainable parameters to trainable functions of the inputs that are calculated instance by instance. As the convolution layer slides across the time dimension, at every time point, it combines relevant information across the multiple time series and surrounding time points into an intermediate features. This layer allows us to capture complex time patterns and cross-variable interactions with shallower networks, providing more effective interpretation compared to traditional CNNs. Next, the results from the convolution layer are combined using an attention layer with the stacked approach shown in Figure 2. This second layer functions in a similar manner to the univariate time series case in Figure 1, combining the intermediate features into the final generated feature.

More formally, for each sample $i$, denote the multivariate time series in Eq (1) as

$$X^{[i]} = \left\{\left\{X^{[i]}_{j,k}\right\}_{k=0:T}\right\}_{j=1:m}. \qquad Eq\ (3)$$

Recall $m$ is the number of time series, and $T$ is the number of time points. The parameter $\tau$ is relatd to the width the convolution layer $\tilde{X}^{[i]}_k = \left\{\left\{X^{[i]}_{j,l}\right\}_{l=(k-\tau):(k+\tau)}\right\}_{j=1:m}$. More specifically, the time window ranges from $k - \tau$ to $k + \tau$, so the width is $(2 * \tau + 1)$. In practice, $\tau$ can be a hyper-parameter that is tuned. For each time point $k$, compute intermediate feature $\Xi^{(i)} = \{\xi^{(i)}_k\}$ as

$$\xi^{(i)}_k = \sum_{j=1}^{m} \sum_{l=-\tau}^{+\tau} A^{(1)}_{j,l}\left(\tilde{X}^{[i]}_k\right) s^{(1)}_{j,l} X^{[i]}_{j,(k+l)}. \qquad Eq\ (4)$$

Here $s^{(1)}_{j,l}$ is the trainable scaling coefficients, and $A^{(1)}_{j,l}\left(\tilde{X}_k\right)$ reflects of the importance of variable $j$ at time $k + l$ for the intermediate feature $\xi^{[i]}_k$. For out-of-boundary scripts of the time windows at the left and right ends, simple padding of zero is needed. For simplicity of notation, we will omit the superscript $[i]$ for $X^{[i]}, \tilde{X}^{[i]}_k$ and $\Xi^{(i)}$ from now on, and denote $\xi^{(i)}_k$ as $\xi_k$, $\tilde{X}^{[i]}_k$ as $\tilde{X}_k$, and $X^{[i]}_{j,k}$ as $X_{j,k}$.

The intermediate feature generating layer in Eq (4) is similar to a regular one-dimensional convolution layer since scaling coefficients $s^{(1)}_{j,l}$ and the parameters constructing $A^{(1)}_{j,l}\left(\tilde{X}_k\right)$ are shift-invariant across different focal time point $j$. However, the selection weights calculated from $A^{(1)}_{j,l}\left(\tilde{X}_k\right)$ are not fixed but constructed as attention subnets (see Section 2.1), and enhance the representing efficiency. We refer to these structures as Convolutional Attention Layer, and the shift-invariant networks $A^{(1)}_{j,l}(\cdot)$ is referred as Kernel Subnets.

Finally, we use another single time-series attention layer to combine all the intermediate features $\{\xi_k\}_{k=1...T}$ into the final generated feature for the sample:

$$Feature = \sum_{k=0}^{T} A^{(2)}_k(\Xi) s^{(2)}_k \xi_k. \qquad Eq(5)$$

We insert Eq (4) to Eq (5), and reorganize the order of summation in the following Eq (6). The generated feature can still be represented by a summation of original inputs weighted by varying coefficients, which preserve the additive nature and help simplify the interpretability:

$$Feature = \sum_{k=0}^{T} A^{(2)}_k(\Xi) s^{(2)}_k \left(\sum_{j=1}^{m} \sum_{l=-\tau}^{\tau} A^{(1)}_{j,l}(\tilde{X}_k) s^{(1)}_{j,l} X_{j,(k+l)}\right)$$

$$= \sum_{j=1}^{m} \sum_{k=0}^{T} X_{j,k} \left[ \sum_{l=-\tau}^{+\tau} A_{j,l}^{(1)}(\tilde{X}_{k-l}) s_{j,l}^{(1)} A_{k-l}^{(2)}(\Xi) s_{k-l}^{(2)} \right]$$

$$\equiv \sum_{j=1}^{m} \sum_{k=0}^{T} W_{j,k}(X) X_{j,k}. \qquad Eq(6)$$

The above feature-engineering head summarizes the entire multivariate time series into a condensed representation, and it will serve as the basic building blocks to extract features from the original time series.

## 2.3 Multi-Head Feature Engineering Machine

In practice, Transformers use multiple attention heads to capture information from different representation subspaces at different positions rather than using a single attention head [4]. So, we will train multiple feature engineering heads in parallel in order to extract potentially different features from the time-series predictors.

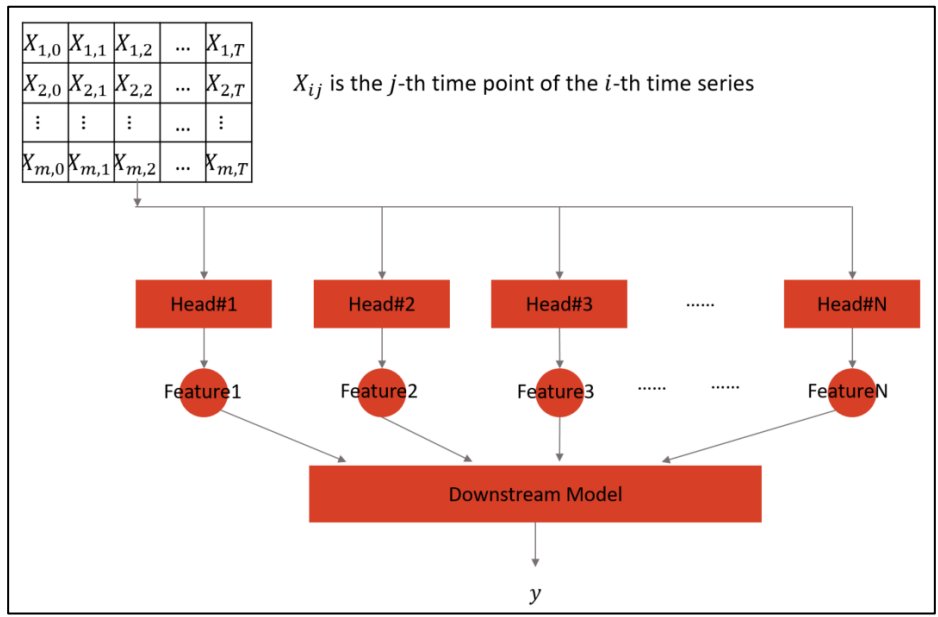

*Figure 3: Multi-head feature engineering machine: Multiple feature engineering heads are trained simultaneously and their generated features are combined in the downstream model*

For our regression problem with multiple time-series predictors, we assume that the true model is

$$Y = F(X) + \epsilon = f_0(f_1(X), f_2(X), \dots, f_n(X)) + \epsilon. \quad Eq\ (7)$$

Our goal is to extract the different features in Eq (7) using the multi-head structure in Figure 3. Each feature engineering head as described in Section 2.2 can be customized to attend on: i) the entire multivariate time series; ii) pre-specified subsets of time-series; or iii) specified time periods.

Depending on the response, we can choose different downstream (DS) models which link the engineered features with final response, such as linear regression models for continuous outcomes and logistic regression models for binary outcomes. The DS model $f_{DS}(\cdot) = f_0(\cdot)$ is part of the feature engineering network and trained simultaneously with the attention heads. Therefore, a relative simple downstream model is preferred to avoid it competing with feature engineering heads. Overly complex downstream models can capture all or part of the interactions, but it is better to let the feature-engineering head capture the non-linearity. Further, a simple downstream model will be more explainable, in terms of indicating which generated features are important. In the rest of the paper, we refer to our approach as feature engineering machine with attention for time series, abbreviated as FEATS.

## 2.4 Hyper-Parameter Tuning

Some hyper-parameters and structural components need to be pre-specified for the proposed architecture, such as (1) the number of feature engineering heads on time series predictors, (2) the attention subnets within feature engineering heads, (3) the width of rolling window for the convolutional attention layer, and (4) the L1 and L2 penalties applied to the scaling parameters $s.^{(\cdot)}$ of each layer. We discuss some guidelines below.

a) A smaller number of feature engineering heads is preferred to produce a parsimonious model. We propose starting with an initial model with a few heads (equal to the dimension of the multivariate time series), then iterate by gradually increasing number of heads (for example, adding two more heads each attempt) and checking if the model performance improves. If this does not meaningfully improve performance, the current number of heads is likely sufficient. Having extra (redundant) heads usually does not harm model performance.

b) It is recommended that the attention subnets within the attention heads should be kept small in size. For example, for a problem with $mT$ around 200, shallow networks with two hidden layers and 10 nodes each layer with ReLU activation are usually sufficient. One of the benefits of using the attention subnets is that neural networks with a parsimonious configuration can reach as good expressivity as large and deep networks [11]. Small ReLU neural network is also more interpretable since the fitted models are locally linear [13].

c) The best choice of hyper-parameter $\tau$, related to the width of rolling windows for the convolutional attention layer in Eq (4), depends on the length of the dependence over time and across the multiple time series [6]. If $\tau$ is larger than the length of dependence across time, the training will force weights for positions near the edge of the kernel subnets will shrink to 0. We recommend using a grid search strategy to select $\tau$ to be somewhere between a small value and 20% of $T$, the length of the time series.

d) Other hyper-parameters are L1 and L2 penalties applied to the scaling parameters $s.^{(\cdot)}$ of different layers. These values are set to 0 by default, but can be tuned via grid search to bring extra sparsity on the selection of variables, time points, and generated features.

## 2.5 Visualization and Explainability of the Attention Heads

We now illustrate how visualization of attention heads can be used to aid in explainability. We take a toy example that consists of three independent time series,

$$X_1 = \{X_{1,k}\}_{k=0:9}, X_2 = \{X_{2,k}\}_{k=0:9}, X_3 = \{X_{3,k}\}_{k=0:9}$$

simulated independently and identically from $N(0,1)$. We then generated the response via

$$y = f_1(X_1, X_2, X_3) + f_2(X_1, X_2, X_3)$$

where

$$f_1(X_1, X_2, X_3) = \frac{1}{3}\left[\max(X_{1,6}, X_{2,6}) + \max(X_{1,7}, X_{2,7}) + \max(X_{1,8}, X_{2,8})\right],$$

$$f_2(X_1, X_2, X_3) = \frac{1}{3}(X_{3,1} + X_{3,2} + X_{3,3}).$$

The exact form of these function is not of particular interest except for the fact that $f_2(\cdot)$ provides a simple linear feature of one time series $(X_3)$, while $f_1(\cdot)$ provides a non-linear interaction of two time series $(X_1, X_2)$, and the two functions are orthogonal.

We then built a multi-head feature engineering algorithm with two attention heads described in Eq (6). The width of convolutional attention layers was set as 0, so the layers focused on selecting subsets of time series attended by heads. The resulting feature engineering machine was used to generate the summaries that follow.

### 2.5.1 Visualizing Attention Weights

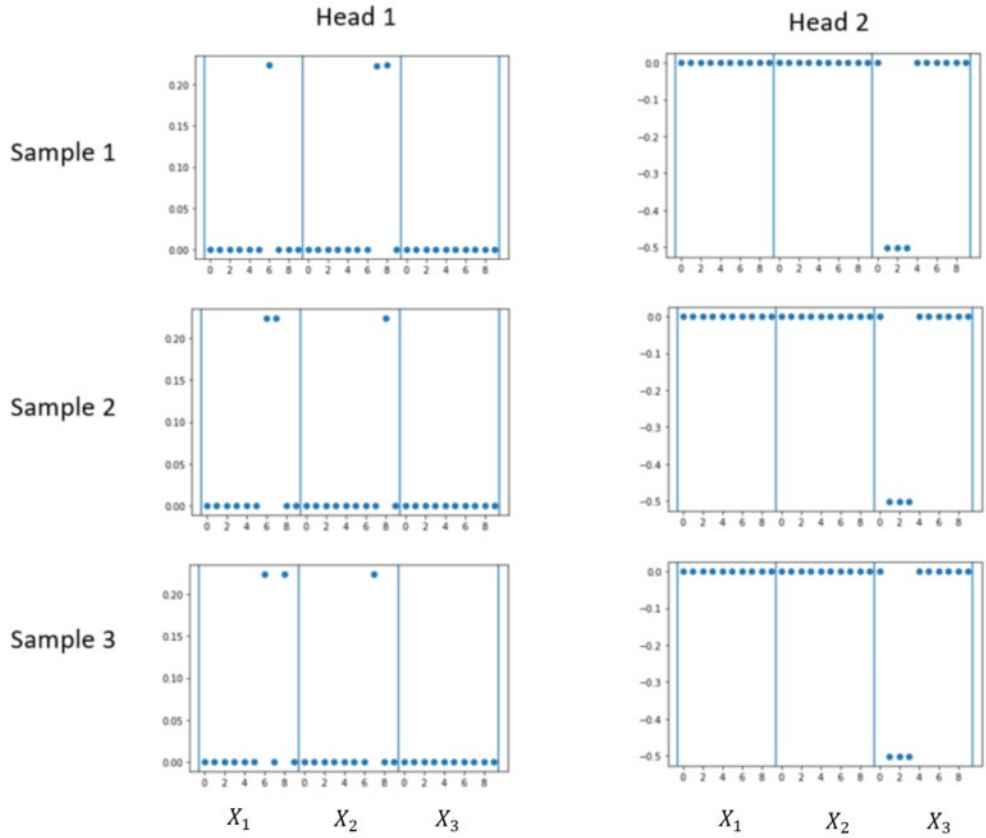

Figure 4: The attention weights $W_{j,k}(X)$ of Head1 and Head2. Each column is plotted for focal attention heads. Each row is plotted for three random samples from the dataset. In each plot, the scatter points are the attention weights $W_{j,k}(X)$ with values on y-axis, and $i, j$ labelled on x-axis.

According to Eq (6), each feature generated from feature engineering head can be expressed as $\sum_{j=1}^{m}\sum_{k=0}^{T}W_{j,k}(X)X_{j,k}$. So, visualizing the varying attention weights $W_{j,k}(X)$ can help understand the feature generating process. Figure 4 shows the weights $W_{j,k}(X)$ of Head1 and Head2 for three samples. The left column shows the weights $W_{j,k}(X)$ of Head1 as a scatter plot. For each, the three different panels within the figure correspond, respectively to $X_{1,0:9}$, $X_{2,0:9}$, $X_{3,0:9}$. We see that, for sample 1, $Feature_1$ can be interpreted as

$$Feature_1 \propto (X_{1,6} + X_{2,7} + X_{2,8}).$$

Similarly, for sample 2,

$$Feature_1 \propto (X_{1,6} + X_{1,7} + X_{2,8}).$$

Finally for sample 3,

$$Feature_1 \propto (X_{1,6} + X_{2,7} + X_{1,8}).$$

Note that the weights for $\{X_3\}$ are all zero. Thus, this feature is aligned with

$$f_1(X_1, X_2, X_3) = \frac{1}{3}\left[\max(X_{1,6}, X_{2,6}) + \max(X_{1,7}, X_{2,7}) + \max(X_{1,8}, X_{2,8})\right].$$

The right column illustrates the attention weights $W_{ij}(X)$ of Head2. For this, across all samples, only the last panel corresponding to $X_{3,0:9}$ have non-zero values, and they are almost the same for the three samples. We see that this feature is aligned with $f_2(X_1, X_2, X_3) = \frac{1}{3}(X_{3,1} + X_{3,2} + X_{3,3})$.

This toy example illustrates the usefulness of visualizing attention weights. In real applications, one can plot the attention weights of multiple samples in an analogous manner to determine whether the pattern is consistent across samples rather than plotting sample by sample. Usually, with noisy data, the interpretation will not be as clean as the toy example, so that additional quantitative assessment is needed. This is the topic of the next section.

### 2.5.2 Quantifying Variable/Time Impacts

Given the additive nature of the generated feature $\sum_{j=1}^{m} \sum_{k=0}^{T} W_{j,k}(X) X_{j,k}$, we can evaluate the contributions of the different time points or time series by comparing the features with the parts of each time point or time series.

- Within each generated feature, for each time series $x_{j,\cdot}$, the directly related internal component is $\sum_{k=0}^{T} W_{j,k}(X) X_{j,k}$
- Within each generated feature, for each time point $x_{\cdot,k}$, the directly related internal component is $\sum_{j=1}^{m} W_{j,k}(X) X_{j,k}$

By calculating and comparing the variance of the generated feature and the variance of its internal components $\sum_{k=0}^{T} W_{j,k}(X) X_{j,k}, j = 1 \ldots m$ and $\sum_{j=1}^{m} W_{j,k}(X) X_{j,k}, k = 0 \ldots T$, we can quantify the influence of the variables and time points to the variability of the generated feature. Table 1 shows the variance comparison for the toy example. We can clearly see which time series and time points are important to the related generated features.

Table 1: The variance of the generated features and their internal components of variables/time points

| | Feature | $x_{1,\cdot}$ | $x_{2,\cdot}$ | $x_{3,\cdot}$ | $x_{\cdot,0}$ | $x_{\cdot,1}$ | $x_{\cdot,2}$ | $x_{\cdot,3}$ | $x_{\cdot,4}$ | $x_{\cdot,5}$ | $x_{\cdot,6}$ | $x_{\cdot,7}$ | $x_{\cdot,8}$ | $x_{\cdot,9}$ |
|---|---|---|---|---|---|---|---|---|---|---|---|---|---|---|
| Head1 | 0.101 | 0.063 | 0.061 | 0 | 0 | 0 | 0 | 0 | 0 | 0 | 0.034 | 0.033 | 0.034 | 0 |
| Head2 | 0.728 | 0 | 0 | 0.728 | 0 | 0.024 | 0.025 | 0.025 | 0 | 0 | 0 | 0 | 0 | 0 |

In addition, Figure 5 shows boxplots of the generated features $\sum_{j=1}^{m}\sum_{k=0}^{T}W_{j,k}(X)X_{j,k}$ for different heads and their internal components calculated for each sample. The left column is for Head1. We can see that the generated feature is closely aligned with $X_3$ and across time points 1,2 and 3, since they are the source of variability of the feature across all samples according to the comparison of boxplots. The right column is for Head2. This feature is aligned with $X_2$ and $X_3$ with almost equal contribution, and across time point 6,7 and 8. This explanation approach is similar to the SHAP's summary plot for average attribution that provide a sense of the distribution of contribution per feature, but needs significantly less post-hoc computation [14].

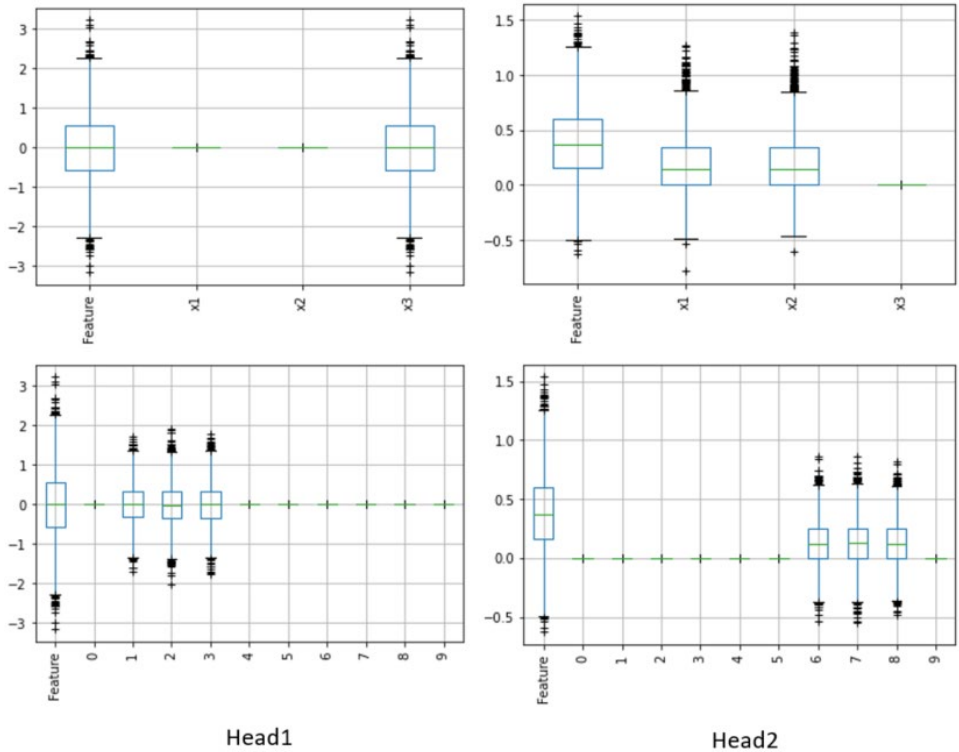

Head1      Head2

Figure 5: Boxplots of generated features and their comparison with internal parts. The y-axis is the generated feature's value of the focal feature engineering head. The x-axis labels which the box represents, the entire feature ("Feature"), or the part related to variable ("$x_1$","$x_2$","$x_3$" for the upper row plots) or time point ("0"-"9" for the lower row plots), The left column is plotted for explaining Head1, the right column is plotted for explaining Head2.

# 3. Simulation Studies

## 3.1 Time Series Predictors, Continuous Outcome

This section uses a simulated dataset with continuous response to illustrate the performance and interpretability of our methodology. The dataset had 55,000 samples that were split into 50,000 for training and 5,000 for testing. Each sample consisted of two time series $X_1^{[i]} = \{X_{1,k}^{[i]}\}_{k=0:49}$ and $X_2^{[i]} = \{X_{2,k}^{[i]}\}_{k=0:49}$ as predictors. They were simulated from two independent heteroscedastic error processes ARCH(1). The outcomes were simulated using the model

$$y_i = 0.005 * \left(X_{1,10}^{[i]} + 3X_{1,11}^{[i]} + 5X_{1,12}^{[i]} + 3X_{1,13}^{[i]} + X_{1,14}^{[i]} - X_{1,15}^{[i]} - 3X_{1,16}^{[i]} - 5X_{1,17}^{[i]} - 3X_{1,18}^{[i]} - X_{1,19}^{[i]}\right)$$
$$+ 0.5 * \max\left(X_{1,30:34}^{[i]}\right) + avg\left(\min(X_{1,k}^{[i]}, X_{2,k}^{[i]})\right)_{k=42:46} + 0.1\, \epsilon_i$$

where $\epsilon_i \sim N(0,1)$. The true model can be decomposed by three features: a linear weighted sum of $X_1^{[i]}$, a non-linear maximum term of $X_1^{[i]}$ and a complex interaction term between $X_1^{[i]}$ and $X_2^{[i]}$. These components overlap across time series, and it will be hard to manually identify and engineer these features.

We implemented our FEATS algorithm with three feature engineering heads. For each head, the width of the convolutional attention layer $\tau$ was set at 3. The attention neural networks were selected as shallow networks with two hidden layers and 10 nodes for each layer. We used ReLU activation with no L1 and L2 penalization. (Unless otherwise specified, these same hyper-parameters were used across all the experiments in Section 3.) For the continuous outcome, we used a simple linear model as the downstream model.

Table 2 shows the performance metrics (MSEs). FEATS algorithm has better performance compared to XGB and FFNN (with 2 hidden layers, 40 nodes each layer). MSE of FEATS is close to 0.10 which is the variance of the noise in the true model.

*Table 2: Performance on simulated dataset*

|  | XGB | FFNN | FEATS |
| --- | --- | --- | --- |
| MSE on Training Dataset | 0.013 | 0.0105 | 0.010 |
| MSE on validation Dataset | 0.0105 | 0.0120 | 0.0108 |

*Table 3: Correlation between the ground truth component and feature extracted from feature engineering heads*

| Correlation | Feature of Head 1 | Feature of Head 2 | Feature of Head 3 |
| --- | --- | --- | --- |
| the weighted sum of $X_1^{[i]}$ | **0.996** | 0.025 | 0.013 |
| $avg\left(\min(X_{1,k}^{[i]}, X_{2,k}^{[i]})\right)_{k=42:46}$ | 0.012 | 0.010 | **0.999** |
| $\max\left(X_{1,30:34}^{[i]}\right)$ | 0.014 | **0.997** | 0.010 |

We see from Table 3 that the feature generated by Head1 is strongly correlated with the linear weighted sum of $X_1^{[i]}$, that for Head2 is strongly correlated with $\max\left(X_{1,30:34}^{[i]}\right)$, and the feature for Head3 is strongly correlated with $avg\left(\min(X_{1,k}^{[i]}, X_{2,k}^{[i]})\right)_{k=42:46}$. The generated features in this example have distinct separation. In real applications, the features are likely to be more correlated.

As noted earlier, the results from the FEATS algorithm can also be interpreted by applying the visualization and explanation approaches in Section 2.5. The first row of Figure 6 is similar to Figure 4, but it shows 50 randomly selected samples $W_{j,k}(X)$ of the focal head stacked on the same panel. For Head1, the curves of different samples are overlapping with each other, which means that the selection on variables and time points consistently give the same weights to the same variables and time points. The pattern represents the specific linear combination from the data-generating model. For Head2, the weights are different across samples with spikes for variable $X_1^{[i]}$ from time point 30 to 34, which align with $\max\left(X_{1,30:34}^{[i]}\right)$. For Head3, the weights are different across sample with spikes for variable $X_1^{[i]}$ and $X_2^{[i]}$ from time point 42 to 46. The weights of pair $X_1^{[i]}$ and $X_2^{[i]}$ are positive and add up to a constant number for each $i$. The pattern represents the complex non-linear interaction $avg\left(\min(X_{1,k}^{[i]}, X_{2,k}^{[i]})\right)_{k=42:46}$.

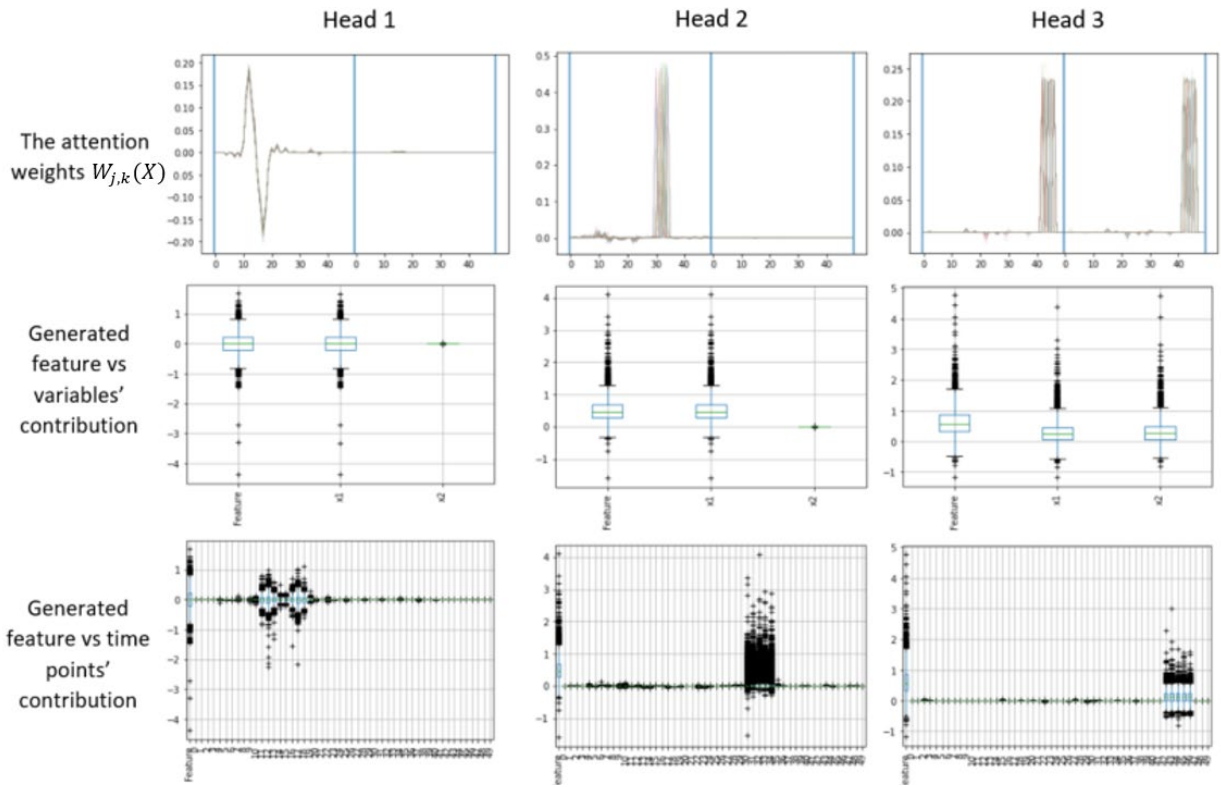

Figure 6: Feature explainability and visualization of the feature engineering heads. Each column is plotted for a focal feature engineering head. For each head, the first row is the stacked attention weights plot of 50 random selected samples, the second and thrid row is the boxplots of generated feature and its comparison with every internal part.

The second and third rows of the plots show the comparison of generated features with the components of the time series and time points internally. The feature of Head 1 is constructed by $X_1^{[i]}$ variable from time 10 to 19; the feature of Head2 is constructed by $X_1^{[i]}$ variable from time 30 to 34; and the feature of Head3 is constructed by both $X_1^{[i]}$ and $X_2^{[i]}$ with equal contribution from time 42 to 46. These observations are aligned with our finding in Table 3, and explain how the feature engineering heads recover the data generating model.

## 3.2 Time Series Predictors, Binary Outcome

This simulation illustrates the performance and interpretability of our methodology on data with a binary outcome. Again, the dataset has 50,000 training samples, and 5,000 testing samples. Each sample $i$ consists of two time series $X_1^{[i]} = \{X_{1,k}^{[i]}\}_{k=0:49}$ and $X_2^{[i]} = \{X_{2,k}^{[i]}\}_{k=0:49}$ as predictors. They are generated from two independent heteroscedastic error processes ARCH(1).

The binary outcome $y$ is simulated from Bernoulli distribution with ground truth log-odds equal to

$$C * \left[ 0.005 * \left( X_{1,10}^{[i]} + 3X_{1,11}^{[i]} + 5X_{1,12}^{[i]} + 3X_{1,13}^{[i]} + X_{1,14}^{[i]} - X_{1,15}^{[i]} - 3X_{1,16}^{[i]} - 5X_{1,17}^{[i]} - 3X_{1,18}^{[i]} - X_{1,19}^{[i]} \right) + 0.5 * \max\left(X_{1,30:34}^{[i]}\right) + avg\left(\min(X_{1,k}^{[i]}, X_{2,k}^{[i]})\right)_{k=42:46} \right],$$

where $C$ is a constant multiplier to tune for different levels of signal-to-noise ratios. We simulated and tested the above datasets three times, with C as 50, 5 and 1 to represent low, medium and high noise. The hyper-parameter configuration is almost the same as in Section 3.1, except using a logistic regression as the downstream model for binary outcome.

Table 4 shows the performance of the FEATS algorithm and its comparison to the oracle and benchmark models. The oracle is calculated assuming the ground truth is known, so it marks the best case for the performance metrics.

*Table 4: Simulation results for time-series as predictors, binary outcome: the accuracy (Accu) and AUC are calculated on the test datasets*

|  |  | Oracle | | XGBoost | | FFNN | | FEATS | |
|---|---|---|---|---|---|---|---|---|---|
| Experiments | $C$ | Accu | AUC | Accu | AUC | Accu | AUC | Accu | AUC |
| Experiment 1 | 50 | 98.46% | 0.9991 | 93.98% | 0.9861 | 96.04% | 0.9912 | 98.16% | 0.9990 |
| Experiment 2 | 5 | 84.76% | 0.9297 | 82.18% | 0.9053 | 83.08% | 0.9147 | 83.94% | 0.9268 |
| Experiment 3 | 1 | 62.30% | 0.6738 | 58.48% | 0.6189 | 60.26% | 0.6422 | 61.10% | 0.6664 |

The FEATS algorithm has better performance on the simulated dataset compared to the benchmark models of XGB and FFNN (with 2 hidden layers, 50 nodes each layer with ReLU activation). Although the performance of FEATS decreases as with the signas-to-noise ratio, its gap with the oracle remains small. In addition, the generated features from the heads can be mapped back to the features in the data generating model with high correlation across all three experiments, with a results similar to those shown in Table 3. Finally, we note that in both Sections 3.1 and 3.2, the FFNN algorithms have 5721 parameters compared to 4684 parameters for FEATS which has better performance.

### 3.3 Static Covariates
### 3.3.1 Modeling

Thus far, we have restricted attention to time-series predictors. In practice, we will have a combination of time-varying and static predictors, so we discuss how to incorporate static covariates into the modeling in this section.

Since our goal is to have interpretable feature engineering, we will use generalized additive models (GAMs) in the initial stage (see Figure 4). Specifically, we fit the GAM structure

$$G(\mathbf{z}) = g_1(z_1) + g_2(z_2) + \cdots + g_p(z_p),$$

using structured neural networks (called GAM-Nets [15]). In this case, the ridge functions $\{g_j(\cdot)\}$ are modeled using sub-networks with only one-dimensional inputs $\{z_k\}$.

For the regression problem with multiple time series and static covariate as predictors, we consider the underlying model to be of the form

$$Y = f(Z, X) + \epsilon = f_{DS}\left(g_1(Z_1), \ldots, g_p(Z_p), f_1(X), f_2(X), \ldots, f_n(X)\right) + \epsilon, \qquad Eq\ (8)$$

where the response depends on transformed static covariates $g_1(Z_1), \ldots, g_p(Z_p)$ and generated dynamic predictors $f_1(X), f_2(X), \ldots, f_n(X)$. To capture the interactions between the time series and static predictors, we adapt the attention layer for a single time series or vector discussed in Section 2.1 to a feature attention layer of the form

$$\hat{y} = \Sigma_j A_j(O) s_j O_j,$$

where $O = \{O_j\}_{j=1:(n+p)} = \{g_1(Z_1), \ldots, g_p(Z_p), f_1(X), f_2(X), \ldots, f_n(X)\}$ and $A_{1\ldots N}(O)$ are calculated from trainable sub-networks. Instead of softmax, we now use the sigmoid activation function

$$A_j(O) = \frac{1}{1 + \exp(-e_j(O))}. \qquad Eq\ (9)$$

The sigmoid activation function in Eq (9) makes the selection of a specific feature independent of the selection of others. The above feature attention layer can also be viewed as a case of multiplicative interactions [11], which is similar to the gating mechanism that is crucial to recurrent neural networks such as LSTMs [16]. Figure 7 shows the architecture with multi-head feature engineering for time-series predictors, GAM-Nets on static covariates, and a feature attention layer for the downstream model. The proposed architecture is quite flexible, with a choice of different building blocks for dynamic feature engineering, static feature engineering, and the downstream model. All these components can be trained simultaneously as parts of FEATS algorithm.

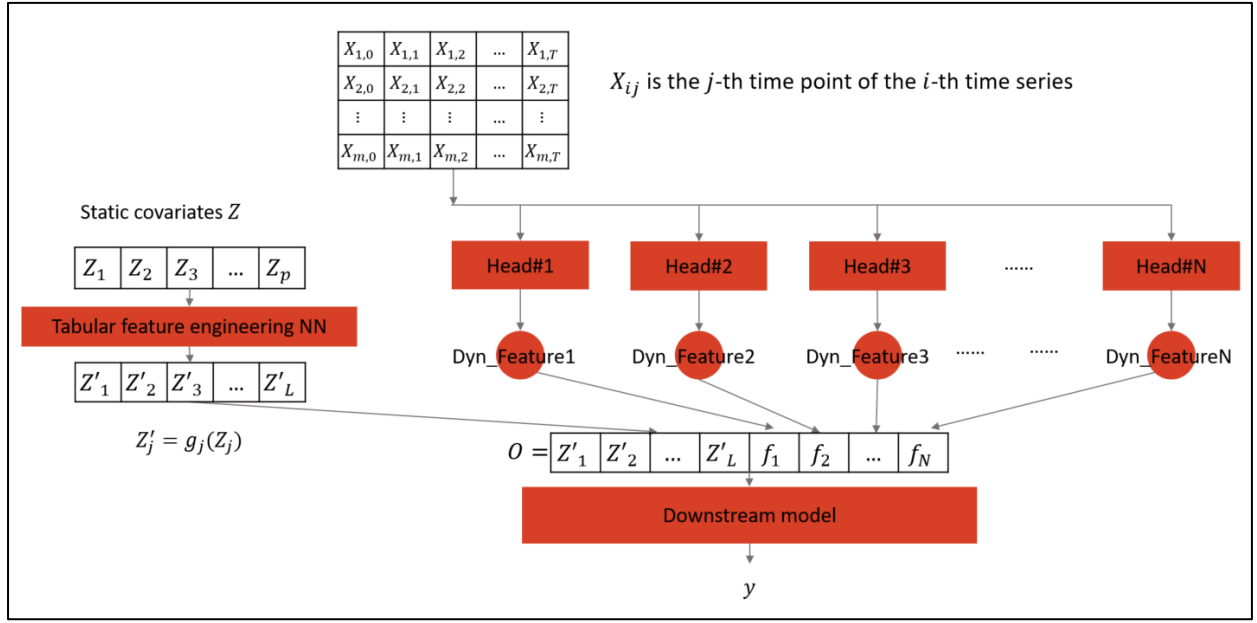

*Figure 7: Feature engineering machine with both multivariate time series and static covariates. The three important structural components for this example implementation are (i) the GAMnets transforming the static covariates, (ii) the multiple attention heads generating features from time series predictors, and (iii) a feature attention layer as the downstream model.*

### 3.3.2 Simulation Study

The dataset has 10,000 training samples and 500 test samples. Each sample $i$ consist of two time series $X_1^{[i]} = \{X_{1,k}^{[i]}\}_{k=0:49}$ and $X_2^{[i]} = \{X_{2,k}^{[i]}\}_{k=0:49}$ independently simulated from ARIMA (p=2,T=1,q=2) processes, which are non-stationary time series. Each sample $i$ also consists of two static covariates $Z_1^{[i]} \sim$ Bernoulli(0.5) and $Z_2^{[i]} \sim N(0,1)$. The model is given by:

$$y_i = avg\left(X_{1,k}^{[i]}\right)_{k=0:10} + Z_1^{[i]} * |Z_2^{[i]}| * avg\left(max\left(X_{1,k}^{[i]}, X_{2,k}^{[i]}\right)\right)_{k=30:35} + Z_1^{[i]}.$$

We fitted the FEATS algorithm, described Figure 7, consisting of three structural components: (i) two feature engineering heads for time series predictors, (ii) GAM-Nets with ridge function $g_1(\cdot)$ for $Z_1^{[i]}$ and $g_2(\cdot)$ for $Z_2^{[i]}$, and (iii) a feature attention layer to combine the generated features and transformed static covariates.

The results are as follows. The mean square error is 0.0009 for the training dataset and 0.0016 for the validation dataset. These values are close to the true variance. Figure 8 shows that the ridge functions of GAM-Nets effectively capture the transformation of the static covariates.

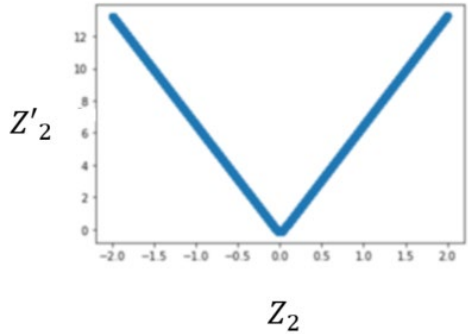

Figure 8: Ridge function $Z'_2 = g_2(Z_2)$

# 4 Real applications

## 4.1 UEA&UCR time series classification repository

We compared the performance of our FEATS algorithm with several others on publicly available datasets (see UEA&UCR time series [18] classification repository for these datasets). The background of the datasets is summarized in Table 5.

Table 5: Time-series classification datasets characteristics

| Dataset | Sample size (train/test) | Dimension of multivariate time-series | Length of time-series | Number of classes |
|---|---|---|---|---|
| BasicMotions | 40/40 | 6 | 100 | 4 |
| HandMovementDirection | 160/74 | 10 | 400 | 4 |
| PenDigits | 7494/3498 | 2 | 8 | 10 |
| StandWalkJump | 12/15 | 4 | 2500 | 3 |
| ArticularyWordRecognition | 278/300 | 9 | 144 | 25 |

We compared our algorithm with a variety of other "benchmark" techniques, including the latest deep learning frameworks [17, 18], the bag-of-patterns model based multivariate time series classification approach [19], and common distance-based classifiers [20]. The techniques we considered for comparison are:

1) TapNet [17]: The attentional prototypical network for multivariate time series classification.
2) MLSTM-FCN [18]: The latest general deep learning framework for multivariate time series classification, consist of an LSTM layer and stacked CNN layer along with a Squeeze-and-Excitation block to generate latent features.
3) WEASEL-MUSE [19]: The latest bag-of-pattern based framework for multivariate time series classification.
4) ED-1NN [20]: One nearest neighbor classifier with Euclidean distance, which is the most popular baseline used for time series classification.

5) 1NN-DTW-D [20]: A variation of the nearest neighbor classifier computes distances based on the Dynamic Time Warping of multi-dimension points.

*Table 6: Accuracy on validation datasets*

| Dataset | FEATS | TapNet | MLSTM-FCN | WEASEL+MUSE | ED-1NN | DTW-1NN-D |
|---|---|---|---|---|---|---|
| BasicMotions | **1** | 1 | 0.95 | 1 | 0.675 | 0.975 |
| HandMovementDirection | **0.4054** | 0.378 | 0.365 | 0.365 | 0.279 | 0.231 |
| PenDigits | 0.968 | 0.98 | 0.978 | 0.948 | 0.973 | 0.977 |
| StandWalkJump | **0.533** | 0.4 | 0.067 | 0.333 | 0.2 | 0.2 |
| ArticularyWordRecognition | 0.963 | 0.987 | 0.973 | 0.99 | 0.97 | 0.987 |

For dealing with multi-class regression we selected a linear dense layer with softmax activation as the downstream model to link with multinomial outcomes. The label with highest score after softmax activation was selected as the prediction. Table 6 lists the performance of the different algorithms on validation datasets. We see that FEATS achieved almost as good or better performance on all the datasets compared to the benchmark approaches.

### 4.2 Trading Dataset

This application deals with high frequency predictions of the market based on streaming tick data from a book service. The direction of the mid-price change in next 3 *ms* could be: i) no change (=0), ii) up (=1), or iii) down (=2). The predictors are multivariate time series consisting of 17 dynamic variables computed online from streaming tick data. They represent the current value and the 10 previous ticks, such as current top of order book bid/ask size, current spread on order book. The length of time series is 11. The training data sample size is 352010, the validation data sample size is 72400, and the test data sample size is 162903.

The FEATS algorithm used 10 heads to extract features from the 17 dynamic variables at the 11 different time points. It used a linear dense layer with softmax activation as the downstream model. The benchmark models were XGBoost using the snapshot time data only (XGB1), XGBoost using all the data (XGB2), Long short-term memory model (LSTM), Generalized additive model network (GAM-Net), and Explainable Neural Network (XNN). Hyper-parameters of all the models were on validation dataset. Performance measures on training and test datasets are listed in Table 7. Since the outcome has three different categories, AUC was calculated as the focal category against the others, and cross entropy loss of multi-class regression is provided.

Only XGB2 achieved slightly better performance than FEATS on the test dataset. But it has a larger loss-gap between the training and test datasets indicating less robustness. Also, FEATS has better model explainability than the XGBoost benchmarks.

Table 7: Model performance on WF Trading Dataset

| AUC (One vs. Others) | Train AUC | | | Test AUC | | | Cross Entropy | |
|---|---|---|---|---|---|---|---|---|
| | 0 (no change) | 1 (increase) | 2 (decrease) | 0 (no change) | 1 (increase) | 2 (decrease) | Train | Test |
| XGB1 | 0.924045 | 0.97996 | 0.979614 | 0.916172 | 0.97945 | 0.978548 | 0.3279 | 0.3347 |
| XGB2 | 0.944287 | 0.985366 | 0.985042 | 0.923558 | 0.981078 | 0.980518 | 0.2862 | 0.3205 |
| LSTM | 0.923012 | 0.979095 | 0.978634 | 0.919857 | 0.980144 | 0.979396 | 0.3329 | 0.3298 |
| GAMnet | 0.923167 | 0.978915 | 0.978846 | 0.920274 | 0.980388 | 0.979758 | 0.3349 | 0.3306 |
| XNN | 0.926773 | 0.980532 | 0.980803 | 0.919623 | 0.980465 | 0.980698 | 0.3246 | 0.3287 |
| TS-Feat | 0.926549 | 0.980497 | 0.980190 | 0.922103 | 0.980905 | 0.979738 | 0.3221 | 0.3283 |

The generated features are quite interpretable. By applying the approaches in Section 2.5, we can explain the results shown in Figure 9 as follows. The lags.BID_SIZE1 is the driven variable of the feature, and time 2 (2 ticks before current) influence more than other time points.

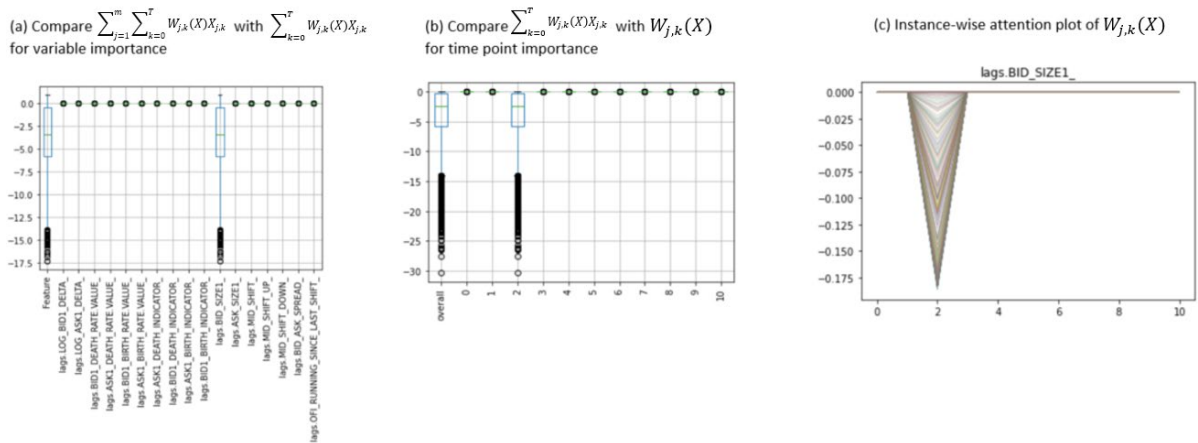

Figure 9: Interpretation plots of a selected feature

### 4.3 Deposit Balance Dataset

The goal here is to provide forecasts of the deposit balances on a monthly periodicity given exogenous hypothetical macroeconomic scenarios. The dataset had five static variables: account time by month, by snapshot time of the forecast, the account balance tier of the pool and two other indicators of the pool's account characteristics. The three multivariate time series are exogenous macroeconomic variables including Federal Funds Rate, GDP, and the ratio of Federal Balance Sheet to GDP. The time series are observed monthly for the current and past 12 months. Samples are weighted by a function of balance tier and pool size. The training sample size is 14,360, the validation sample size is 2,000, and the test sample size is 2,000. The target was to predict the log-transformed balance change of the next month at the snapshot time for the sample pool.

The FEATS algorithm used six feature engineering heads to extract the features from the multivariate time series at the 13 time points, used GAM-Net with 2 ridge functions for each of the five static variables, and a feature attention layer described in Section 2.3 to combine the generated features and transformed static covariates. The benchmark models were XGBoost using the static variables only (XGB1), XGBoost using the static variables and the dynamic variables at the snapshot time only (XGB2), XGBoost using the static variables and the dynamic variables across all time. Hyper-parameters of all models were tuned on validation dataset. The performances on training and test dataset are given in Table 8. We see that FEATS is better than the benchmark models.

*Table 8: Model performance on WF Deposit Balance Dataset*

| Model | Train $R^2$ score | Test $R^2$ score |
|---|---|---|
| XGB1 | 0.7005 | 0.5771 |
| XGB2 | 0.7250 | 0.6253 |
| XGB3 | 0.7814 | 0.7719 |
| FEATS | **0.7946** | **0.7823** |

The FEATS algorithm found that among the static variables, the MOB (months on book) that record the age of account is the most influential variable. The longer history an account has, the more positive the balance change in the next month. Further, the percentage of change will be larger for accounts with more than 16 months history. For the multivariate time series, the top important features were mostly driven by GDP and the Federal Fund Rate. These findings are aligned with subject-matter knowledge.

## 5. Conclusion

We have proposed attention networks for interpretable feature engineering of time-series predictors. Our multi-head feature engineering algorithm (FEATS) extracts the underlying features from multivariate time-series predictors accurately. In addition, the additive nature of the networks means the extracted features are more interpretable than traditional networks. The multi-head feature engineering machine can be used as a primary model or a part of a more complex network, or its primary goal can be feature engineering. Regardless, the attention heads provide very good explanations of how the features summarize complex multivariate time series.